\documentclass{article}
\usepackage{spconf,amsmath,graphicx}
\usepackage{bm}
\usepackage{amsmath}
\usepackage{amssymb}
\usepackage{multirow}
\usepackage{float}
\usepackage{dblfloatfix}
\usepackage{mathrsfs}
\usepackage{enumitem}
\usepackage{graphicx}
\usepackage{subfigure}
\usepackage{booktabs} % for professional tables
\usepackage{hyperref}
\usepackage[utf8]{inputenc} % allow utf-8 input
\usepackage[T1]{fontenc}    % use 8-bit T1 fonts
\usepackage[square,numbers]{natbib}
\usepackage{hyperref}       % hyperlinks
\usepackage{url}            % simple URL typesetting
\usepackage{booktabs}       % professional-quality tables
\usepackage{amsfonts}       % blackboard math symbols
\usepackage{nicefrac}       % compact symbols for 1/2, etc.
\usepackage{microtype}      % microtypography
\usepackage[dvipsnames]{xcolor}
\usepackage{bbm}
\usepackage{enumitem}
\usepackage[title]{appendix}
\usepackage{caption}
\usepackage{subfigure}
\usepackage{fixltx2e}
\usepackage{eqnarray}

\title{fMRI Data augmentation via Synthesis}

\name{$Peiye \ Zhuang^1$ \hspace{11pt} $Alexander\ G.\ Schwing^2$ \hspace{11pt} $Oluwasanmi\ Koyejo^{1,3}$}

\address{
University of Illinois at Urbana-Champaign\\
$\ ^1$Dept. of Computer Science, $\ ^2$Dept. of Electrical and Computer Engineering,$\ ^3$Beckman Institute\\
\{peiye, aschwing, sanmi\! \}@illinois.edu}

\begin{document}
%\ninept

\maketitle

%\textcolor{red}{How to give an appealing title?}
\begin{abstract}
%Understanding the individual variability of brain function and its association with behavior is one of the major concerns in modern cognitive neuroscience. To this end, 
%With limited sample sizes and high-dimensional structured samples, data augmentation is considered a promising approach for improving the quality of complex fMRI predictive models. 
%%to understand how the common fMRI data shortage can be addressed. 
We present an empirical evaluation of fMRI data augmentation via synthesis. For synthesis we use generative models trained on real neuroimaging data to produce novel task-dependent functional brain images. Analyzed generative models include classic approaches such as the Gaussian mixture model (GMM), and modern implicit generative models such as the generative adversarial network (GAN) and the variational autoencoder (VAE). In particular, the proposed GAN and VAE models utilize 3-dimensional convolutions, which enables  modeling of high-dimensional brain image tensors with structured spatial correlations. The synthesized datasets are then used to augment classifiers designed to predict cognitive and behavioural outcomes. 
Our results suggest that the proposed models are able to generate high-quality synthetic brain images which are diverse and task-dependent. Perhaps most importantly, the performance improvements of data augmentation via synthesis are shown to be complementary to the choice of the predictive model. Thus, our results suggest that data augmentation via synthesis is a promising approach to address the limited availability of fMRI data, and to improve the quality of  predictive fMRI models.
%As far as we known, we are the first to compare capability of the models generating high quality synthetic brain images.
%which model is able to generate high quality synthetic brain images which are diverse and task dependent. 
%Our model is constructed as a conditional generative adversarial network combined with three dimensional convolutions. 
%This architecture enables the modeling of high dimensional brain image tensors with structured spatial correlations. 
\end{abstract}

\vspace{-0.3cm}
\section{Introduction}
\label{sec:intro}
\vspace{-0.2cm}
%Human brain activity, as measured by functional Magnetic Resonance Imaging (fMRI), varies significantly between individuals. In response, modern analysis now prioritizes understanding the inter-subject variability of brain function ~\citep{dubois2016building}. 
Progress in computational cognitive neuroimaging research is stifled by the difficulty of obtaining large quantities of brain imaging data~\citep{poldrack2014making}. This is especially apparent in {\em decoding} studies where machine learning methods are used to predict cognitive and behavioral outcomes of brain imaging experiments~\citep{varoquaux2014machine}. To this end, research has focused on increasingly sophisticated predictive models that can take advantage of specialized properties of brain images~\citep{cox2003functional, pereira2009machine, nathawani2016neuroscience}.
%But so far, no suitable model addresses pressing data issues in cognitive neuroscience. For the computational neuroscientist, generated images deliver unlimited quantities of high quality brain imaging data that can be used to develop state of the art tools before application to real subjects and/or patients ~\citep{varoquaux2014machine}. 
Our work is motivated by the view that generative models provide another useful tool for improving the performance of brain decoding models via data augmentation, and furthermore, that synthesis-based data augmentation is complementary to advances in classification models.
%-- as they enable the synthesis of a variety of plausible brain images representing different hypothesized individuals.
%and high-quality generative models can be analyzed to posit potential mechanisms that explain this variability ~\citep{van2008individual}. \textcolor{red}{The last two sentences seem duplicated.}
Indeed, data augmentation is a standard approach in computer vision~\citep{Goodfellow-et-al-2016}, where it has been shown to improve the performance of predictive models. Classic data augmentation is focused on linear transformations such as rotation and scaling, and nonlinear methods such as adding noise. 
More recently, %computer vision researchers have begun to explore data augmentation using modern 
implicit generative models have been proposed to capture rich notions of natural variability~\citep{Richardson2018OnGA}. %This research has shown that sophisticated data augmentation via synthesis can lead to improvements in classification performance \textcolor{red}{cite, discussed ref}. 

Inspired by this progress, our work seeks to answer: {\bf {\em Can data  augmentation using generative models be used to improve fMRI classification performance?}} This manuscript provides – to our knowledge for the first time, affirmative results suggesting that it is indeed possible to generate artificial, high-quality, diverse, and task-dependent functional brain images. Furthermore, we provide empirical evidence that the synthesized data can be used for data augumentation -- resulting in improved fMRI classifier performance. To better understand fMRI data augmentation we provide results for Gaussian mixture models (GMMs), variational auto-encoders (VAEs) and generative adversarial nets (GANs). In addition to qualitative evaluation, we show quantitative results demonstrating that classifiers trained using the generated images combined with real images result in improved performance as compared to classifiers trained only with real images.

\begin{figure*}[t]
%\vspace{-1.4cm}
  \centering
  \includegraphics[width=0.97\linewidth]{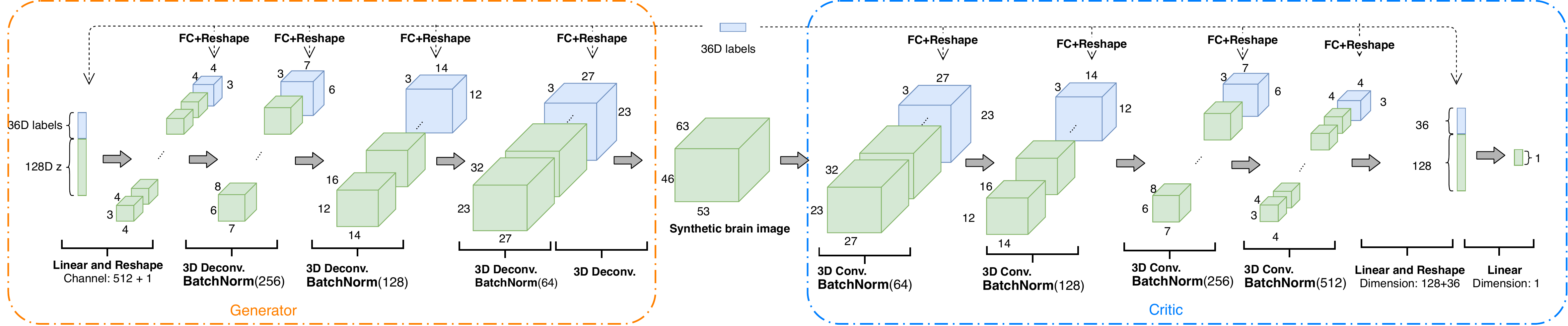}
%  \vspace{-0.5cm}
  \caption{ICW-GAN architecture. For the generator (orange box), the 128-dimensional encoding $z$ is drawn from a multivariate Gaussian. The label vector is a binary encoding. It is concatenated to input and hidden layers, and for each of the 4 layers, fully connected layers followed by a tanh activation transform the label vector to volumes of appropriate size. %Our stride in the de-convolutional layers is [1,2,2,2,1] in the batch, height, width, length and feature map dimension. Batch normalization is leveraged in the feature map dimension, sizes of which are in parentheses, i.e. 64 and 128. 
  The structure of the discriminator (blue box) is a mirrored generator. 
  }
  \label{fig:ciwgan.G}
\vspace{-0.3cm}
\end{figure*}

\vspace{-0.2cm}
\section{Background and Related Work}
\label{relatedworks}
%\vspace{-0.1cm}
%\subsection{Neural Nets for fMRI Brain Data}
\vspace{-0.2cm}
%We are unaware of any published work using neural networks to generate brain imaging data. However, 
Deep neural networks have been used for classifying brain imaging data~\citep{firat2014deep,koyamada2015deep,nathawani2016neuroscience,svanera2017deep}. Our approach differs in that we will use generative models to obtain artificial fMRI data. We assume the reader to be familiar with GMMs and review only briefly VAEs and GANs in the following. %and~\citep{koyamada2015deep}, used 2D deep nets to extract features of fMRI brain images to classify brain states. \citep{nathawani2016neuroscience} applied both 2D and 3D neural networks to classify fMRI brain data. \citep{svanera2017deep} decoded fMRI data of video stimuli and classified data into visual categories. Similarly, authors of ~\citep{nathawani2016neuroscience} extracted features from 4D fMRI data and used deep learning methods for discrimination of cognitive processes. 

\noindent\textbf{Variational auto-encoders (VAEs)}: VAEs aim to maximize the parametric likelihood $p_\theta(x)$, where $\bm{x}$ is a sample from a dataset. %The encoding step in VAEs is to encode inputs to a latent vector $z$ with a prior distribution $p_\theta(z)$, while the decoding step is to reconstruct the latent vector $z$ back to the input space that will be as similar as the original input $x$. Because the posterior inference $p_\theta(x|z)$ is intractable for VAE, a lower bound is used as an objective function, which can be written as
To this end, a low-dimensional space represented by variable $\bm{z}$ is hypothesized and one optimizes the lower bound
\begin{eqnarray}
%\begin{split}
\log p_\theta(\bm{x}) \geq -\text{KL}(q_\phi(\bm{z}|\bm{x}), p(\bm{z})) + \nonumber \\ \mathbb{E}_{q_\phi(\bm{z}|\bm{x})} [\log p_\theta(\bm{x}|\bm{z})], \nonumber
\label{eq:vae}
%\end{split}
\end{eqnarray} 
which compares the approximate posterior $q_\phi$ to the prior $p(\bm{z})$ using the $\text{KL}$-divergence, while simultaneously estimating reconstruction via the decoding $p_\theta(\bm{x}|\bm{z})$. 

\noindent\textbf{Generative adversarial nets (GANs):} 
To learn a distribution over data $\bm{x}$, a GAN~\citep{goodfellow2014generative} formulates a 2-player non-cooperative game between two deep nets: (1) the generator $G$ which depends on a random noise vector $\bm{z}$ sampled from a known prior  distribution $P_{\bm z}$  and produces an image $G_\theta(\bm{z})$; % The generator $G$ is trained to fool 
(2) the discriminator $D$ which receives  synthetic  or real data, and is trained to differentiate as accurately as possible. Since the generator is tasked to make differentiation as hard as possible the resulting formulation is a saddle-point objective. 
% Formally, $G$ and $D$ play the following two-player minimax game with value function $V(G,D)$:
% %\vspace{-0.3cm}
% \begin{equation}
%     \min_{G}\max_{D}V(G,D):=\mathbb E_{\bm x \sim P_{data}(\bm x)}[\log D(\bm x)] + \mathbb E_{\bm z\sim P_{\bm z}(\bm z)}[\log(1-D(G(\bm z)))].
% \end{equation}
% \label{eq:gan}
% %\vspace{-0.2cm}
%In a conditional GAN~\citep{mirza2014conditional}, both $G$ and $D$ are conditioned on some extra information $\bm{y}$, for instance, class labels or other features. This conditioning can be accomplished by feeding an encoding of $\bm{y}$ into both the generator and the discriminator. This approach is successful in one-to-many mappings such as image labeling with many tags.

%The Wasserstein GAN (WGAN)~\citep{arjovsky2017wasserstein} objective uses the \emph{Wasserstein-1} distance:
%\begin{equation}
%\label{eq:wgan}
%   \min_\theta \max_{w \in \emph{W}}\mathbb E_{\bm x \sim P_{data}(\bm x)}[D_w(\bm x)] - \mathbb E_{\bm z\sim P_{\bm z}(\bm z)}[D_w((G_\theta(\bm z)))],
%\end{equation}
%where $\{D_w\}_{w \in \emph{W}}$ denotes a set of functions  that are  $K$-Lipschitz for some $K$. 
%Intuitively, the Wasserstein metric between distributions measures the minimum cost of transporting mass to transform the distribution $P_{data}$ into the distribution $P_{\bm z}$. This loss is continuous everywhere and its gradient with respect to its input has been found to be more stable than its classical GAN counterpart. 

Many formulations have been proposed to improve upon the original idea~\citep{arjovsky2017wasserstein}. Here, we use the Improved Wasserstein GAN (IW-GAN) framework~\citep{gulrajani2017improved}  % that weight clipping of the critic in WGANs inevitably causes the gradient to either vanish or to explode. To address this issue, ~\citep{gulrajani2017improved} proposes an alternative penalty term in the critic loss based on the gradient norm when optimizing:
\iffalse
\begin{equation}
    \mathbb E_{\bm z\sim P_{\bm z}(\bm z)}[(1-D_w(G_\theta(\bm z)))]-\mathbb E_{\bm x \sim P_{data}(\bm x)}[D_w(\bm x)] +  \lambda \mathbb E_{\bm \hat x \sim P_{\bm{\hat x}}}[(\parallel { \nabla_{\bm {\hat x }} D_w(\bm {\hat x}) \parallel }_2 -1 )^2 ].
    \label{eq:IWGAN}
\end{equation}
\fi
which optimizes w.r.t.\ parameters  $\theta$ and $w$
\begin{eqnarray}
%\begin{split}
\mathbb E_{\bm z\sim P_{\bm z}}[D_w(G_\theta(\bm z))]-\mathbb E_{\bm x \sim P_{r}}[D_w(\bm x)] \\\nonumber 
+\ \lambda \mathbb E_{\bm{\hat x} \sim P_{\bm{\hat x}}}[(\parallel { \nabla_{\bm {\hat x }} D_w(\bm {\hat x}) \parallel }_2 -1 )^2 ],
\label{eq:IWGAN}
\vspace{-0.2cm}
%\end{split}
\end{eqnarray}
where $P_r$ is the real data distribution, $\bm{\hat x}$ is a convex combination of real data and artificial samples, i.e., ${\bm {\hat x}} \gets \epsilon \bm x + (1- \epsilon)G_\theta(\bm z)$ with $\epsilon$ drawn from a uniform distribution (${\epsilon \sim \emph{U}[0,1]}$), and ${\lambda}$ is a gradient penalty coefficient.

%{\color{red}DELETE? [Sanmi: Agreed] In an auxiliary classifier GAN (AC-GAN)~\citep{odena2016conditional}, every generated sample has a corresponding label which can be used to condition the \emph{generator} and the \emph{classifier}. Further, the \emph{discriminator} is modified to contain an auxiliary decoder network to reconstruct the training data.} % Formally, the objective function  differs from other versions of GANs, which also contain the log-likelihood of the classifier.
3D-GANs~\citep{wu2016learning} extend GANs to 3D object generation. Different from classical GANs, 3D-GANs apply 3-dimensional convolutions in both the generator and the discriminator. By learning deep object representations, 3D GANs can generate visually appealing yet variable 3D object volumes. We use conditional variants of both VAEs and GANs~\citep{cvae, mirza2014conditional}.

%\input{sections/method}
%\vspace{-0.3cm}
%\input{sections/experiment}
%\vspace{-0.2cm}
%\input{sections/conclusion}
%%%%%%%%%%%%%%
%%%%%%%%%%%%%%
%%%%%%%%%%%%%%
%\subsection{Variational auto-encoders (VAE)}

%,where $\phi$ the parameters of the encoder and $\theta$ the parameters of the decoder.

%\textbf{Conditional Variational auto-encoder (CVAE)}:
%When it comes to data with multi-classes, a one-to-many mapping can be an efficient tool in modeling the conditional distribution $p(x|c)$ where $x$ is a data and $y$ is condition, i.e., a label. Then the lower bound can be extended as follows:

%\begin{eqnarray}
%\begin{split}
%log(p_\theta(x|y)) &\geq -KL(q_\phi(z|x, y)|| p_\theta(z)) + \\\nonumber 
%&\mathbb{E}_{q_\phi(z|x, y)} [-log(p_\theta(x|z, y))] 
%\label{eq:IWGAN}
%\end{split}
%\end{eqnarray}

%The practical loss for CVAE can be written as 
%\begin{eqnarray}
%\begin{split}
%\widetilde{\mathcal{L}}_{CVAE} = &-KL(q_\phi(z|x, y)|| p_\theta(z)) + \\\nonumber 
%&\frac{1}{L} \sum\limits^{L}_{l=1} p_\theta(y|x,z^{(l)})
%\label{eq:IWGAN}
%\end{split}
%\end{eqnarray}
%, where $L$ is the number of samples.
\vspace{-0.4cm}
\section{Approach}
\label{app}
\vspace{-0.3cm}

We use a GMM to directly model the high-dimensional data likelihood, using the expectation-maximization (EM) algorithm to learn parameters of the Gaussian distributions from training data. We slightly adjust conditional VAEs (CVAEs) and GANs as discussed subsequently: for CVAEs, we use 3-dimensional convolutions which enable modeling of the spatial relation within high-dimensional brain images. We modify GANs in a similar way and discuss our adjustments for an improved conditional Wasserstein GAN (ICW-GAN) briefly. %In the following, we introduce a 3D Improved Conditional Wasserstein GAN (ICW-GAN) model for fMRI data generation, different from existing GAN models in structure and use of label information. 

Similar to classical GANs, ICW-GANs are formulated as a non-cooperative two-player game between two adversaries: a generator $G_\theta$ and a discriminator $D_w$.
%(1) a generator $G_\theta(z)$, which generates artificial samples $\hat{\bm x} $ from randomly drawn latent encodings $z$ via a transformation using a deep net parameterized by $\theta$; and (2) a discriminator $D_w(\bm x)$  %which computes the probability that a given data point $x$ is real rather than artificial, by using  represented via the logit %$f_w(\bm x)$  obtained from a deep net parameterized by $w$. %$D_w(x) = \sigma(f_w(x))$ that a given sample 

Different from classical GANs, 3D convolution and deconvolution are used to capture the spatial structure of the voxel information. Moreover, both the discriminator $D_w$ and the generator $G_\theta$ are conditioned on available labels. We also use the Wasserstein distance and gradient penalty~\citep{gulrajani2017improved} in the objective function. As a result, our ICW-GAN integrates 3D-GANs, conditional GANs and IW-GANs.

%We construct our models to minimize the improved Wasserstein distance in Equation \ref{eq:IWGAN}. Our model extends the IW-GAN in two ways. First, because fMRI data is three dimensional, 
%Further, in order to stabilize training, the propsed ICW-GAN applies a more stable upper bound for Wasserstein distance as an objective function, compared to previous work, 
%{\color{red}{Could you expand more on this?}}
%\textcolor{red}{maybe compress/cut off this paragraph?}
The overall model architecture is illustrated in Fig.~\ref{fig:ciwgan.G}. Our generator consists of 4 fully convolutional layers, % with kernels of size ${4\times 4\times 4}$ and stride 2, 
batch normalization, ReLU layers added in between, and a sigmoid layer at the end. The discriminator architecture is a mirrored generator except for the final layer which uses a linear activation.
To include label information, we concatenate labels to the input and hidden layers. %At the input of the generator, binary labels are combined with the brain vector. Then, for each of the intermediate layers, we use a fully connected layer followed by a $\tanh$ activation to transform the binary vector to a volume of appropriate size, i.e., ${3\times4\times3}$ for the first hidden layer, and ${3\times7\times6}$ for the next. We empirically found that the model is not sensitive to the choice of volume size. We concatenate the label volume to intermediate volumes and pass the joint to the next deconvolutional layer. We follow the same procedure in the architecture of the discriminator.
We note in passing that we experimented with concatenating labels in various layers and did not observe significant differences. Thus, we chose to implement the model using full concatenation of labels.

\noindent\textbf{The objective function} of the ICW-GAN model is % as follows:
\vspace{-0.2cm}
\begin{equation}
\label{eq:Loss1}
\begin{split}
L=&\mathbb E_{\bm z\sim P_{\bm z}}[(D_w(G_\theta(\bm z| \bm y)))]-\mathbb E_{\bm x \sim P_{r}}[D_w(\bm x|\bm y)]\\
&+\ \lambda \mathbb E_{\bm {\hat x} \sim P_{\bm{\hat x}}}[(\parallel { \nabla_{\bm {\hat x }} D_w(\bm {\hat x| \bm y}) \parallel }_2 -1 )^2 ].\nonumber
\end{split}
\end{equation}
%\vspace{-0.8cm}
\vspace{-0.1cm}

\noindent\textbf{Downstream Classifiers
%\footnote{Random forest and Spacenet classifiers are also tried. Their not so well performance on `Real' and `Real+Synth' is still under discussion.}
}
%however (several tenths decrease in accuracy rate)
\label{sec:classifier}
%\vspace{-0.2cm}
We consider classification with data augmentation to quantitatively investigate the hypothesis that the generated images accurately reproduce the conditional image statistics.
%To quantitatively evaluate the generative model, we consider the task of enlarging the dataset for downstream tasks such as classification using synthetically generated samples. 
%We employ a variety of downstream classifiers and evaluation configurations. Common classifiers in neuroscience are Support Vector Machines (SVMs) and 3D deep networks. 
We note that support vector machines (SVMs) and deep neural net classifiers are state of the art for fMRI applications~\citep{pereira2009machine} and we consider it sufficient to employ them for evaluation. 
To this end, we compare the SVM and the 3D deep net classifier trained with real brain images (`Real') or real plus synthetic brain images (`Real+Synth.'). Accuracy, macro F1, precision, and recall metrics are used to measure the results.
%, i.e., about a dozen parameters for SVMs and several hundred thousand for our deep networks. which gives us some insight into both linear and non-linear classification performance. W % obtained from different training datasets.

\begin{table}[t]
\centering
%\begin{minipage}{0.48\linewidth}
\setlength{\tabcolsep}{0.5pt}
% Downsampling reflects the down-scaling of brain volumes. 
\caption{Classification results for collection 1952. %`Input' represents the input of a classifier: `Real' indicates using only real training data; `Real+noise' indicates using real data plus real data adding Gaussian noise; `Real+Synth' indicates real training data plus synthetic data which is produced by the ICW-GAN. `Gen. model' means the generative model while `-' indicates no generative model is used. Two classifiers are explored: SVM and neural networks (DNN). Results show that augmenting real data with synthetic data improves classification performance.
}
\vspace{-0.2cm}
{\footnotesize
\begin{tabular}{cc|ccccc}
\toprule
\textbf{Input}& \textbf{Gen. model} & \textbf{Classifier} & \textbf{Accuracy} & \textbf{Macro F1} & \textbf{Precision} & \textbf{Recall}\\
\midrule 
Real &- &SVM &0.8181 &0.82 & 0.8333 & 0.8133 \\
Real+noise &- &SVM &0.8185 &0.82 &0.8367 &0.82 \\
Real+Synth.	&GMM &SVM &0.8188 &0.82 &0.8366	&0.82 \\

Real+Synth.	&CVAE &SVM &0.8248	&0.8267	&0.8367	&0.8233 \\

Real+Synth.	&ICW-GAN &SVM &\textbf{0.8311}	&\textbf{0.83}	&\textbf{0.8433}  &\textbf{0.8333} \\
\midrule 
Real &- &DNN &0.852 &0.857 & 0.872 & 0.8523 \\
Real+noise &- &DNN &0.8581 &0.856 &0.8719	&0.8579 \\
Real+Synth.	&GMM &DNN &0.8604 &0.8631 &0.8749 &0.8604 \\
Real+Synth.	&CVAE &DNN &0.8684 & 0.869 &0.8827 &0.8683 \\
Real+Synth.	&ICW-GAN &DNN &\textbf{0.8799} &\textbf{0.8825} &\textbf{0.8933} &\textbf{0.88} \\

%\midrule 
%Real &SVM &0.855 &0.857 &0.867 &0.857 \\
%Real &DNN & 0.863 &0.863 &0.872 &0.863 \\
%Real+Synth.  &SVM &0.860 &0.863 &0.860 &0.857 \\
%Real+Synth.  &DNN &\textbf{0.891} &\textbf{0.894} &\textbf{0.906} &\textbf{0.891}\\
\bottomrule
\end{tabular}
}
%\vspace{-0.2cm}
\label{tab:Exp1}
\vspace{-0.3cm}
\end{table}

%\end{minipage}
%\begin{minipage}{0.48\linewidth}
\begin{table}[t]
\centering
\setlength{\tabcolsep}{0.5pt}
% Downsampling reflects the down-scaling of brain volumes. 
\caption{Classification results for collection 2138.}
%\begin{center}
%\setlength{\tabcolsep}{3pt}
\vspace{-0.2cm}
{\footnotesize
\begin{tabular}{cc|ccccc}
\toprule
\textbf{Input}& \textbf{Gen. model} & \textbf{Classifier} & \textbf{Accuracy} & \textbf{Macro F1} & \textbf{Precision} & \textbf{Recall}\\
\midrule 
Real &- &SVM &0.6234 &0.5967 & 0.6	&0.6233 \\

Real+noise &- &SVM &0.6348 &0.6133 &0.61 &0.6367 \\

Real+Synth.	&GMM &SVM &0.6385 &0.6 &0.6033	&0.6333 \\

Real+Synth.	&CVAE &SVM &\textbf{0.6428} &\textbf{0.62} &0.6233 &0.6397 \\
Real+Synth.	&ICW-GAN &SVM &0.6404 &\textbf{0.62} &\textbf{0.6267}  &\textbf{0.64} \\

\midrule 
Real &- &DNN &0.7028 &0.6939 &0.7162	&0.7029 \\
Real+noise &- &DNN &0.7353 &0.723 &0.7373 &0.7353 \\

Real+Synth.	&GMM &DNN &0.7393 &0.723 &0.7303  &0.7393\\

Real+Synth.	&CVAE &DNN &\textbf{0.7503} &0.738 &0.7533  &\textbf{0.7503} \\

Real+Synth.	&ICW-GAN &DNN &0.7494 &\textbf{0.7393} &\textbf{0.759}  &0.7493\\

%\midrule 
%Real &SVM &0.855 &0.857 &0.867 &0.857 \\
%Real &DNN & 0.863 &0.863 &0.872 &0.863 \\
%Real+Synth.  &SVM &0.860 &0.863 &0.860 &0.857 \\
%Real+Synth.  &DNN &\textbf{0.891} &\textbf{0.894} &\textbf{0.906} &\textbf{0.891}\\
\bottomrule
\end{tabular}
}
%\vspace{-0.2cm}
\label{tab:exp2138}
%\end{center}
\vspace{-0.3cm}
%\end{table*}
%\end{minipage}
\end{table}

%We optimize both the discriminator and generator loss using an Adam optimizer~\citep{kingma2014adam}.
\begin{figure}[t]
%\vspace{-1.4cm}
  \centering
  \includegraphics[width=\linewidth]{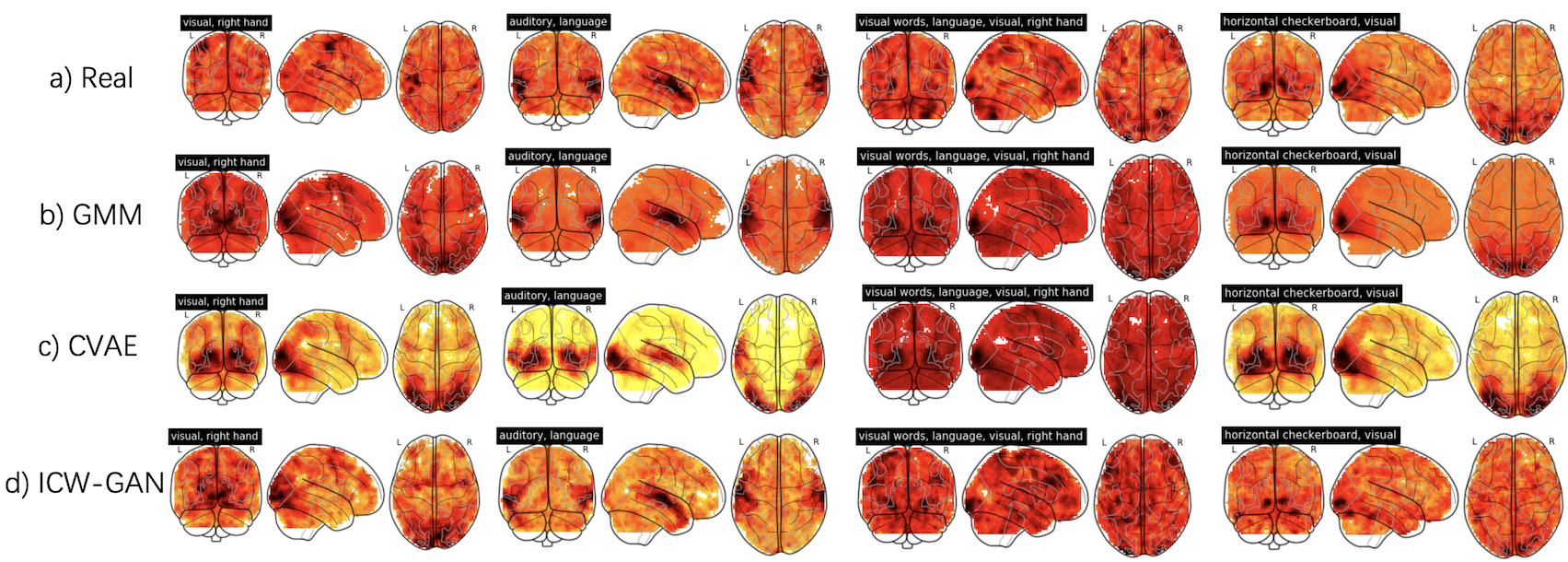}
  \vspace{-0.6cm}
  \caption{Real and synthetic data of collection 1952. % with same classes. 
  Each row shows  brain images, either real or synthetic data generated by GMM, CVAE and ICW-GAN. Each column presents the brain images of the same class. Classes from left to right: `visual, right hand', `auditory, language', `visual words, language, visual, right hand', and `horizontal checkerboard, visual'.}
  \label{fig:generated}
\vspace{-0.5cm}
\end{figure}

\textbf{SVM}: 
As is common for fMRI data, we train a simple linear SVM on masked training data to classify the masked test data. By applying the computed mask to a brain volume, invalid voxels are discarded and valid voxels are placed in a 1-dimensional vector, thus reducing the dimension of the brain volumes. 
%\footnote{the mask applying tool can be found in http://nilearn.github.io/modules/generated/nilearn.masking.apply\_mask.html}. 
The mask is computed based on the training data. Several strategies of mask computation can be used, e.g., computing the mask corresponding to gray matter part of the brain or computing the mask of the background from the image border. We conduct the experiments on several masking strategies and do not find much difference.  %We set the value of C in the SVM 1.0 and used hinge loss. 
%\textcolor{red}{Any papers can be referenced to say that by using extracted features, SVM can have better performance?} 

%We use a simple linear SVM to classify test data and do not extract any intermediate features. Instead, as is common in whole brain classification literature~\cite{pereira2009machine, varoquaux2014machine}, we use raw brain data, vectorized to a 1-dimensional vector. 
% for classification to avoid any form of collusion. % result of SVM comparable to that of a neural network classifier which might be better to leverage raw data rather than extracted features. 
%When using SVM, the first step is to transform a brain volume to a 1-D vector. Then, we use them as the input of SVM and classify 1-D transformed vectors.

\textbf{Deep Neural Net}: The deep neural net structure is similar to the discriminator with a 3-dimensional structure and an identical number of convolution layers with Leaky ReLU activations. Unlike the discriminator, the classifier obviously does not concatenate intermediate and input data with any label information.
%Moreover, instead of only discriminating between real or synthetic brain volumes, generally, the classifier has a higher dimensional output depending on the number of classes of the dataset. 

\vspace{-0.5cm}
\section{Data Augmentation Analysis}
\label{experiment}
\vspace{-0.3cm}
%A specialized cross validation strategy is used in all reported results to assess the classifier performance. 
%We also show the training loss curve of the ICW-GAN to illustrate the training stability of our model. 
%\textcolor{red}{Check again to in case of ignoring some details}

In this section we quantitatively and qualitatively analyze the proposed techniques, shedding light on data augmentation for fMRI data.  
First, we present quantitative results for 3D volume classification via training of downstream classifiers on mixtures of real and synthetic data. We further employ as data augmentation methods a Gaussian Mixture Model (GMM), a CVAE, and an ICW-GAN. %We use and use downstream classifiers to compare their generating performance on brain images. 
A stratified 3-fold cross validation is leveraged for all experiments which means that we maintain the percentage of brain images for each class in each fold. %Our cross-validation is slightly different from the classic cross-validation strategy. 
%To avoid test data leakage, we split test data to 3 folds and mixed up the two folds with training data and tested on the remained fold. We repeated the experiments until each fold of testing data being evaluated. The visual illustration of such cross-validation strategy is in our appendix. 
%Beyond that, 
We conduct each classification experiment three times and average the results in order to improve statistical reliability. We will also present detailed qualitative results via examples of generated 3D volumes.

 % \caption{\textwidth 5cm we only use generated data from the ICW-GAN to train the deep net classifier. Two light brown lines mean only using real data with the two classifiers.}
 %This figure shows the trajectory of multi-class accuracy along with the number of artificial training samples resembling collection 1952 at the low resolution setup. The horizontal axis represents the number of artificial data used for each class, while the vertical axis represents the accuracy value. In this case,  note the accuracy obtained by
 
Neurovault~\citep{gorgolewski2015neurovault} is currently the largest open database of preprocessed neuroimaging data (particularly focused on cognitive neuroscience). We evaluate the performance of the  generative methods on the two large Neurovault functional brain image collections 1952 and 2138 which are publicly available. %\footnote{Collections are publicly available online, e.g., https://neurovault.org/collections/2138}
%.\textcolor{red}{``The final PDF should not have any links or bookmarks.'' -- compliance of ISBI.} 
In the preprocessing step, we apply min-max normalization on the brain images to a 0-1 range.
%\vspace{-0.3cm}
%\subsection{Results}
%\vspace{-0.1cm}

\noindent\textbf{Dataset 1 (Collection 1952) results:}   
Collection 1952 is obtained from OpenfMRI, the Human Connectome Project, and the Neurospin research center. The dimensions of the brain images in  collection 1952 are $53 \times 63 \times 46$. Containing 6573 brain images and 45 classes with a total number of 19 sub-classes, e.g., a multi-label encoding, collection 1952 is designed to map a wide set of cognitive functions. The labels are the set of cognitive processes associated with the image, e.g., `visual,' `language,' and `calculate.' %A combination of 2 to 4 sub-classes are subsumed for a class. 
Over the dataset, classes with more than than 100 images are split into training, validation and test subsets with a ratio 7:1:2. For classes with less than 100 images but more than 30 images, we use a 3:1:2 data split. Nine classes with less than 30 samples are ignored. This leaves 4375, 675, and 1398 training, validation, and test brain images, including a total of 36 classes with at least 30 examples. A one-hot encoding is applied to the 36 classes.

%Note that we ensure that the test data always consists of real images only. 
We employ a data augmentation method and the generative models mentioned above on collection 1952. Recall that the original brain images are normalized to the 0-1 range. Thus we choose to test Gaussian noise with a mean value of 0 and a variance of 0.01 as data augmentation. We also experimented with variances of 0.01, 0.05, 0.1 and 0.3 and observed that augmentation performed  best with a variance of 0.01 which we refer to as `Real+noise.' % Thus in the following, we show our classification results with the variance of 0.01.%\textcolor{red}{<-- Can we say like this without presents results with var ='other values'?} 
We train a separate Gaussian Model for each label. %The priors of GMMs are adjusted according to the data. 
When synthesizing a new image, we first choose a specific class and then randomly sample from the trained GMM. We train the CVAE and the ICW-GAN with a batch size of 50, optimized by an Adam Optimizer with an initial learning rate of 1e-4 until convergence. The dimension of the latent variables in the CVAE is 128. Only one CVAE and one ICW-GAN model are trained for collection 1952.

To further assess the quality of the generated data, we evaluate the performance of downstream classifiers. To train the classifiers we use either real data, real data plus 100 real data with additive Gaussian noise per class, or real data plus 100 generated data per class.
Note that the test data for classification is always composed of real images. The classification results are shown in Table~\ref{tab:Exp1}. The first column indicates the type of  training data we use for the classifier. %: only using real data (`Real'), using real data and real data adding Gaussian noise (`Real+noise'), or using the mixed data of real and generated volumes (`Real+Synth'). 
The second column represents the generative source of the synthetic data. The third column denotes the classifier type, i.e., an SVM or a deep neural net (`DNN'). We use the validation dataset to choose the best training models and use these models to classify the test data. We observe that including the synthetic data is generally beneficial for classifier training.
% We also observe that the deep net classifier generally outperforms SVMs.

\noindent\textbf{Dataset 2 (Collection 2138) results:}  
Collection 2138  includes data from the Individual Brain Charting (IBC) project, developed to collect high resolution fMRI of 12 subjects that undergo a large number of tasks: the HCP tasks, the ARCHI tasks, a specific language task, video watching, low-level visual stimulation, etc. 
There are 1847 brain images, 61 classes and 50 labels in collection 2138.  The labels are encoded with a one-hot scheme. Because of the small size of the dataset, we randomly choose 70\% of the brain images as training data and leave 30\% as test data. 
%In this case, we do not have development data to supervise the training process. Thus, we train our models in several runs {\color{blue}what exactly does this mean?} and record the best 
The dimension of brain images in collection 2138 is $ 105 \times 126 \times 91$, which is relatively large; thus we downsample\footnote{Downsampling and parcellation are common practices in fMRI analysis, as standard preprocessing renders fMRI images to be spatially smooth~\citep{arslan2017human}. In addition, there is significant evidence in the fMRI literature that downsampling has limited effect on classifier performance (particularly cross-subject brain alignment and subsequent smoothing, see Figure 3 of~\citep{poldrack2013toward}). 
This is not necessarily true for other kinds of brain imaging. For example, structural brain images can be reliably analyzed at higher resolution.} the brain images to  ${53\times63\times46}$. % using the nilearn python package\footnote{http://nilearn.github.io}. 
Similar to the experiments on collection 1952, we apply simple data augmentation and generative methods, followed by  downstream classifiers. % on collection 2138. 
The dimension of the latent variables in the CVAE is 32. The input of the classifiers can be only real data, real data plus 20 real data with additive Gaussian noise per class, or real data plus 20 generative data per class. The classification results are summarized in Table~\ref{tab:exp2138}. 

Again we observe that usage of synthetic data is generally beneficial for classifier training. Furthermore, the deep neural network generative models -- the CVAE and the ICW-GAN, can significantly outperform the addition of noise and the GMMs for classification. Besides, we also observe that the deep net classifier generally outperforms linear SVMs.

\iffalse
\begin{table*}[b]
% Downsampling reflects the down-scaling of brain volumes. 
\caption{Classification results for the collection 2138.}
\begin{center}
\setlength{\tabcolsep}{3pt}
\begin{tabular}{cc|ccccc}
\toprule
\textbf{Input}& \textbf{Gen. model} & \textbf{Classifier} & \textbf{Accuracy} & \textbf{Macro F1} & \textbf{Precision} & \textbf{Recall}\\
\midrule 
Real &- &SVM &0.6234 &0.5967 & 0.6	&0.6233 \\

Real+noise &- &SVM &0.6348 &0.6133 &0.61 &0.6367 \\

Real+Synth.	&GMM &SVM &0.6385 &0.6 &0.6033	&0.6333 \\

Real+Synth.	&CVAE &SVM &\textbf{0.6428} &\textbf{0.62} &0.6233 &0.6397 \\
Real+Synth.	&ICW-GAN &SVM &0.6404 &\textbf{0.62} &\textbf{0.6267}  &\textbf{0.64} \\

\midrule 
Real &- &DNN &0.7028 &0.6939 &0.7162	&0.7029 \\
Real+noise &- &DNN &0.7353 &0.723 &0.7373 &0.7353 \\

Real+Synth.	&GMM &DNN &0.7393 &0.723 &0.7303  &0.7337\\

Real+Synth.	&CVAE &DNN &\textbf{0.7503} &0.738 &0.7533  &\textbf{0.7503} \\

Real+Synth.	&ICW-GAN &DNN &0.7494 &\textbf{0.7393} &\textbf{0.759}  &0.7493\\

%\midrule 
%Real &SVM &0.855 &0.857 &0.867 &0.857 \\
%Real &DNN & 0.863 &0.863 &0.872 &0.863 \\
%Real+Synth.  &SVM &0.860 &0.863 &0.860 &0.857 \\
%Real+Synth.  &DNN &\textbf{0.891} &\textbf{0.894} &\textbf{0.906} &\textbf{0.891}\\
\bottomrule
\end{tabular}
%\vspace{-0.2cm}
\label{tab:exp2138}
\end{center}
\vspace{-0.5cm}
\end{table*}
\fi

%\textcolor{red}{Should give more analysis about results. e.g. Recall, Precision, }

\begin{figure}[t]
\vspace{-0.3cm}
  \centering
  \includegraphics[width=\linewidth]{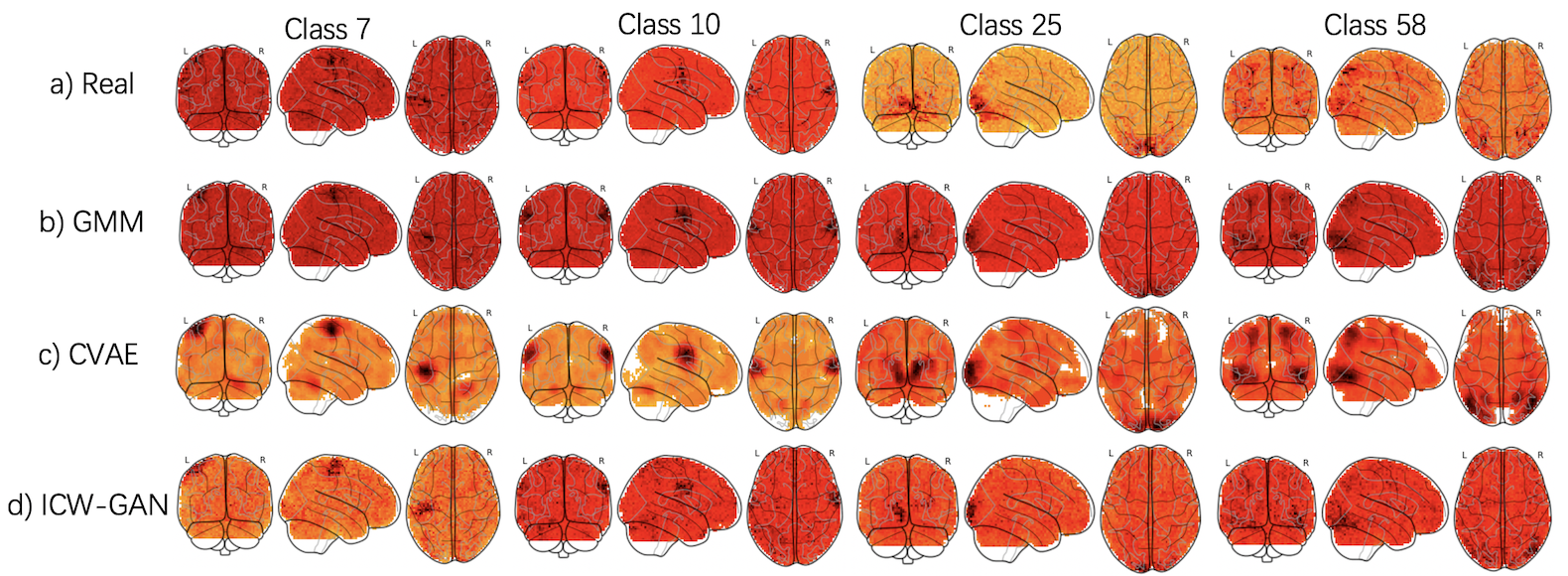}
 \vspace{-0.8cm}
  \caption{Real and synthetic data of collection 2138. % with same classes. 
  Each row shows the brain images from real or synthetic data generated by GMM, CVAE and ICW-GAN. The brain images in each column have the same class label.}
  \label{fig:2138_generated}
\vspace{-0.5cm}
\end{figure}

\noindent\textbf{Visualization of synthetic images}: 2D projections of several real and synthetic brain volumes of  collection 1952 and collection 2138 are illustrated in Fig.~\ref{fig:generated} and Fig.~\ref{fig:2138_generated}. The voxels of the volumes are normalized to $[0,1]$. %, in order to compare all visual results in the same range. 
The dark areas in the projections represent high voxel intensities, i.e., significant activity. 
The projections in Fig.~\ref{fig:generated} are real and synthesized brain images conditioned on the accompanying cognitive process labels shown on the top left. 
The classes in Fig.~\ref{fig:2138_generated} represent types of the tasks received by the subjects  %\textcolor{red}{seems better to only give illustration for one class and leave others in the appendix.}
%\vspace{-0.3cm}
%\begin{itemize}
%\item[$\ast$] 
(Class 7: `response execution', `right finger response execution'; 
%\vspace{-0.3cm}
%\item[$\ast$] 
Class 10: `response execution', `tongue response execution'; 
%\vspace{-0.3cm}
%\item[$\ast$] 
Class 25: `response execution', `working memory', `place maintenance', `visual place recognition'; 
%\vspace{-0.3cm}
%\item[$\ast$] 
Class 58: `working memory', `string maintenance', `visual string recognition'). 
%\end{itemize}
We  observe that the  generative models accurately learn active regions for the both datasets. 
Visual examination of the generated images by neuroscience experts suggests high quality and high diversity. In particular, experts report high activation in the appropriate brain regions, e.g., the motor cortex for motor labels, and the visual cortex for visual labels. 

\begin{table}[t]
\centering
% Downsampling reflects the down-scaling of brain volumes. 
\caption{Variances for 3-fold cross validation on data 1952.}
\setlength{\tabcolsep}{0.5pt}
\vspace{-0.3cm}
{\footnotesize
\begin{tabular}{cc|ccccc}
\toprule
\textbf{Input}& \textbf{Gen. model} & \textbf{Classifier} & \textbf{Accuracy} & \textbf{Macro F1} & \textbf{Precision} & \textbf{Recall}\\
\midrule 
Real &- &SVM &2.8e-4	 &3.0e-4 & 4.3e-4	& 4.3e-4\\
Real+noise &- &SVM &6.8e-5 &1.0e-4 &4.4e-5	&1.0e-4 \\
Real+Synth.	&GMM &SVM &3.1e-4 &4.0e-4 &2.3e-4	&4.0e-4 \\

Real+Synth.	&CVAE &SVM &7.5e-5	&1.3e-4	&1.3e-4 &3.3e-5 \\
Real+Synth.	&ICW-GAN &SVM &2.9e-4	&3.0e-4	&1.3e-4 &4.3e-4 \\
\midrule 
Real &- &DNN &2.8e-4 &3.1e-4 & 3.3e-4	&2.8e-4 \\
Real+noise &- &DNN &2.5e-4 &2.4e-4 &2.4e-4  &2.5e-4 \\
Real+Synth.	&GMM &DNN &3.3e-5 &3.1e-5 &2.2e-5	&3.4e-5 \\

Real+Synth.	&CVAE &DNN &1.5e-5	 &2.3e-6 &1.4e-5 &1.4e-5 \\

Real+Synth.	&ICW-GAN &DNN &1.3e-4 &9.0e-5 &3.0e-5	&1.4e-4 \\

\bottomrule
\end{tabular}
}
%\vspace{-0.2cm}
\label{tab:var1952}
\vspace{-0.5cm}
\end{table}

\iffalse
\textbf{Classification results}: To further assess the quality of the generated data, we evaluate the performance of downstream classifiers. Note that the test data for classification is always composed of real images. The classification results are shown in Table~\ref{tab:Exp1}. The first column indicates the type of  training data we use for the classifier: only using real data (`Real'), or using the mixed data of real and generated volumes (`Real+Synth'). The second column denotes the classifier type, i.e., an SVM or a deep neural network (`DNN'). We use the validation dataset to choose the best training models and use these models to classify the test data. We observe that including the synthetic data is generally beneficial for classifier training. We also observe that the deep net classifier generally outperforms SVMs.
\fi

\noindent\textbf{Variance of cross-validated performance in ICW-GAN:} 
%\label{section:var}
We test our model in various cross-validation settings and calculate the mean variance of the evaluation metrics over the three folds (Table~\ref{tab:var1952}) on collection 1952. The small variances suggest that the reported accuracy differences are indeed statistically significant.

\vspace{-0.3cm}
\section{Conclusion}
\label{conclusions}
\vspace{-0.3cm}
%Generative models provide a useful tool for analyzing brain images. 
The results of this manuscript compare -- to our knowledge for the first time, the performance of different generative models trained on brain imaging data for data augmentation. In our experiments, 3D generative models, particularly the conditional VAE and the ICW-GAN, are shown to generate high quality, diverse, and task dependent brain images. Beyond qualitative evaluation, we evaluate quantitative performance by using the generated images as additional training data in a predictive model -- mixing synthetic and real data to train classifiers. The results show that our synthetic data augmentation can improve classification accuracy. The ICW-GAN and the CVAE easily outperform the generative baselines of GMMs and data augmentation with Gaussian noise, which illustrates that not all data augmentation methods are equally beneficial for the task of image generation. We hope our results inspire additional research on generative models for brain imaging data.
%What should be notified is that, we use downstream classification to exemplify the significance of brain data generation but applications are not only limited to classification tasks. In other words, our models can contribute to many medical cognitive tasks laking sufficient data.  
Future work will focus on additional qualitative evaluation of the generated images by neuroscience experts and exploration of additional applications. We also plan to more thoroughly investigate the trained models to explore what they may contribute to the science of individual variability in neuroimaging. Finally, we plan to expand our models to combine data across multiple studies -- each of which use different labels, by exploring techniques for merging labels based on the underlying cognitive processes~\citep{poldrack2006can}.

\newpage

\bibliographystyle{IEEEbib}
\bibliography{refs}

\begin{thebibliography}{10}

\bibitem{poldrack2014making}
R.~A. Poldrack and K.~J. Gorgolewski,
\newblock ``{Making big data open: data sharing in neuroimaging},''
\newblock {\em Nature neuroscience}, 2014.

\bibitem{varoquaux2014machine}
G.~Varoquaux and B.~Thirion,
\newblock ``{How machine learning is shaping cognitive neuroimaging},''
\newblock {\em GigaScience}, 2014.

\bibitem{cox2003functional}
D.~D Cox and R.-L Savoy,
\newblock ``{Functional magnetic resonance imaging (fMRI)“brain reading”:
  detecting and classifying distributed patterns of fMRI activity in human
  visual cortex},''
\newblock {\em Neuroimage}, 2003.

\bibitem{pereira2009machine}
F.~Pereira, T.~Mitchell, and M.~Botvinick,
\newblock ``{Machine learning classifiers and fMRI: a tutorial overview},''
\newblock {\em Neuroimage}, 2009.

\bibitem{nathawani2016neuroscience}
D.~D. Nathawani, T.~Sharma, and Y.~Yang,
\newblock ``{Neuroscience meets deep learning},'' 2016.

\bibitem{Goodfellow-et-al-2016}
I.~Goodfellow, Y.~Bengio, and A.~Courville,
\newblock {\em Deep Learning},
\newblock MIT Press, 2016.

\bibitem{Richardson2018OnGA}
E.~Richardson and Y.~Weiss,
\newblock ``On gans and gmms,''
\newblock {\em CoRR}, vol. abs/1805.12462, 2018.

\bibitem{firat2014deep}
O.~Firat, L.~Oztekin, and F.~Vural,
\newblock ``{Deep learning for brain decoding},''
\newblock in {\em Image Processing (ICIP), 2014 IEEE International Conference
  on}, 2014.

\bibitem{koyamada2015deep}
S.~Koyamada, Y.~Shikauchi, K.~Nakae, M.~Koyama, and S.~Ishii,
\newblock ``Deep learning of fmri big data: a novel approach to
  subject-transfer decoding,''
\newblock {\em CoRR}, 2015.

\bibitem{svanera2017deep}
M.~Svanera, S.~Benini, G.~Raz, T.~Hendler, R.~Goebel, and G.~Valente,
\newblock ``{Deep driven fMRI decoding of visual categories},''
\newblock {\em arXiv preprint arXiv:1701.02133}, 2017.

\bibitem{goodfellow2014generative}
I.~J. Goodfellow, J.~A. Pouget, M.~Mirza, B.~Xu, D.~Warde-Farley, S.~Ozair,
  A.~Courville, and Y.~Bengio,
\newblock ``{Generative Adversarial Nets},''
\newblock in {\em Advances in neural information processing systems}, 2014.

\bibitem{arjovsky2017wasserstein}
M.~Arjovsky, S.~Chintala, and L.~Bottou,
\newblock ``{Wasserstein GANs},''
\newblock {\em arXiv preprint arXiv:1701.07875}, 2017.

\bibitem{gulrajani2017improved}
I.~Gulrajani, F.~Ahmed, M.~Arjovsky, V.~Dumoulin, and A.~Courville,
\newblock ``{Improved training of Wasserstein GANs},''
\newblock {\em Advances in Neural Information Processing Systems}, 2017.

\bibitem{wu2016learning}
J.~Wu, C.~Zhang, T.~Xue, B.~Freeman, and J.~Tenenbaum,
\newblock ``{Learning a probabilistic latent space of object shapes via 3d
  generative-adversarial modeling},''
\newblock in {\em Advances in Neural Information Processing Systems}, 2016.

\bibitem{cvae}
K.~Sohn, H.~Lee, and X.Yan,
\newblock ``Learning structured output representation using deep conditional
  generative modelss,''
\newblock {\em Advances in Neural Information Processing Systems}, 2015.

\bibitem{mirza2014conditional}
M.~Mirza and S.~Osindero,
\newblock ``{Conditional Generative Adversarial Nets},''
\newblock {\em arXiv preprint arXiv:1411.1784}, 2014.

\bibitem{gorgolewski2015neurovault}
KJ~Gorgolewski, G.~Varoquaux, G.~Rivera, Y.~Schwarz, SS~Ghosh, C.~Maumet, VV.
  Sochat, T.~E-Nichols, Russell~A Poldrack, J-B. Poline, et~al.,
\newblock ``Neurovault. org: a web-based repository for collecting and sharing
  unthresholded statistical maps of the human brain,''
\newblock {\em Frontiers in neuroinformatics}, vol. 9, 2015.

\bibitem{arslan2017human}
S.~Arslan, S.~I. Ktena, A.~Makropoulos, and et.al,
\newblock ``Human brain mapping: a systematic comparison of parcellation
  methods for the human cerebral cortex,''
\newblock {\em NeuroImage}, 2017.

\bibitem{poldrack2013toward}
R.~A. Poldrack, D.~M. Barch, J.~Mitchell, and et.al,
\newblock ``Toward open sharing of task-based fmri data: the openfmri
  project,''
\newblock {\em Frontiers in neuroinformatics}, vol. 7, pp. 12, 2013.

\bibitem{poldrack2006can}
R.~A. Poldrack,
\newblock ``Can cognitive processes be inferred from neuroimaging data?,''
\newblock {\em Trends in cognitive sciences}, 2006.

\end{thebibliography}
%\bibliography{strings,refs}

\end{document}